\documentclass[lettersize,journal]{IEEEtran}
\usepackage{amsmath,amsfonts}
\usepackage{amsthm}
\usepackage{algorithmic}
\usepackage{algorithm}
\usepackage{array}
\usepackage[caption=false,font=normalsize,labelfont=sf,textfont=sf]{subfig}
\usepackage{textcomp}
\usepackage{stfloats}
\usepackage{url}
\usepackage{verbatim}
\usepackage{graphicx}
\usepackage{cite}
\newtheorem{theorem}{Theorem}
\newtheorem{definition}{Definition}
\newtheorem{proposition}{Proposition}
\newtheorem{corollary}{Corollary}
\newtheorem{lemma}{Lemma}
\newtheorem{example}{Example}

\newcommand{\bdfn}{\begin{definition}}
\newcommand{\edfn}{\end{definition}}

\newcommand{\bthm}{\begin{theorem}}
\newcommand{\ethm}{\end{theorem}}
\newcommand{\bprop}{\begin{proposition}}
\newcommand{\eprop}{\end{proposition}}
\newcommand{\blem}{\begin{lemma}}
\newcommand{\elem}{\end{lemma}}

\newcommand{\bcor}{\begin{corollary}}
\newcommand{\ecor}{\end{corollary}}
\newcommand{\bex}{\begin{example}}
\newcommand{\eex}{\end{example}}

\newcommand{\bobs}{\begin{observation}}
\newcommand{\eobs}{\end{observation}}

\begin{document}

\title{Optimisation Is Not What You Need}

\author{\IEEEauthorblockN{Alfredo Ibias,~\IEEEmembership{Member,~IEEE,}\\
\IEEEauthorblockA{Avatar Cognition\\
Barcelona, Spain\\
Email: alfredo@avatarcognition.com}
}
\thanks{Manuscript received June 20, 2025; revised XXX XX, 2025.}}

\markboth{IEEE Transactions on Emerging Topics in Computational Intelligence,~Vol.~XX, No.~X, August~2025}%
{Ibias: Optimisation Is Not What You Need}

\IEEEpubid{0000--0000/00\$00.00~\copyright~2025 IEEE}

\maketitle

\begin{abstract}
The Artificial Intelligence field has focused on developing optimisation methods to solve multiple problems, specifically problems that we thought to be only solvable through cognition. The obtained results have been outstanding, being able to even surpass the Turing Test. However, we have found that these optimisation methods share some fundamental flaws that impede them to become a true artificial cognition. Specifically, the field have identified catastrophic forgetting as a fundamental problem to develop such cognition. This paper formally proves that this problem is inherent to optimisation methods, and as such it will always limit approaches that try to solve the Artificial General Intelligence problem as an optimisation problem. Additionally, it addresses the problem of overfitting and discuss about other smaller problems that optimisation methods pose. Finally, it empirically shows how world-modelling methods avoid suffering from either problem. As a conclusion, the field of Artificial Intelligence needs to look outside the machine learning field to find methods capable of developing an artificial cognition.
\end{abstract}


\section{Introduction}
\IEEEPARstart{T}{here} is a common goal in the Artificial Intelligence field: approaching the achievement of an artificial cognition by producing results similar to those produced by a natural cognition (i.e. a human). That is, the efforts in such field have been focused on mimicking the effects of cognition. This approach has produced a plethora of optimisation methods that try to solve problems that are considered solvable only by humans. From simple pattern matching problems like classifying MNIST images~\cite{deng12}, to more complex problems like the Turing Test~\cite{turing50}. The underlying assumption was that, if some algorithm is able to solve these problems, it will be due to the emergence of cognition (or at least some kind of cognition-like reasoning).

Special importance has the Turing Test example, because it was based on the assumption that only a cognitive agent would be able to process language. This perception came from the observation that language is one of the fundamental developments that differentiate us from other animals, although that is not as true as popularly believed~\cite{ssrr85}. However, the test was arguably passed in $2023$~\cite{biever23}, and somehow the emergence of an artificial cognition has not happened~\cite{smahbf25}. This has been the highest disappointment of the aforementioned approach, while at the same time opening a whole can of worms in the philosophical field.

To explain this lack of emergent cognition even when resolving very complex problems, the scientific community has found some critical caveats in current Artificial Intelligence solutions. This paper focus in the problem of \emph{catastrophic forgetting}, probably the main caveat that explains the non-emergence of cognition. The main consensus today is that this caveat can be overcome by some kind of patch or small modification to current solutions. However, its presence in multiple methods, and the lack of success by the attempts to overcome it, can be a hint that implies that this problem may be more fundamental than it is usually considered to be. Thus, it is necessary to explore it from a formal perspective, with the aim of understanding what makes it so commonplace and how it may be overcome.

\IEEEpubidadjcol
This paper explores the catastrophic forgetting problem as a derivative of the optimisation setup. It proves that, due to the guidance provided by the loss function (either the explicit one or the implicit ones), most machine learning and deep learning methods are unable to overcome the effects of catastrophic forgetting. One could argue about how far such effects could go, and about how much they can be mitigated, but the underlying reality is that they are inevitable. This situation can be intuitively understood due to the weighted nature of these methods: if the whole algorithm is based on tuning some weights to solve a problem, and whenever a new problem comes those weights have to adapt, then they will be modified and lose the capability to solve the initial problem. In fact, that intuition is why most of the proposed solutions to overcome catastrophic forgetting are based on limiting how much (or how many) weights can change. In the end, the inevitability of catastrophic forgetting is a road-block in the path towards Artificial General Intelligence using these methods, and hence novel approaches are necessary.

Additionally, this paper explores the overfitting problem and how it also impedes the development of an artificial cognition. In this case it is already widely accepted that current machine learning and deep learning methods will always suffer from overfitting. Thus, this paper only shows that this situation is also associated with the fact that these methods are developed with an optimisation setup in mind. Moreover, the paper discuss about how this is a problem for developing Artificial General Intelligence, and how little sense this problem has from a natural cognition perspective.

To add a positive contribution, this paper also shows empirically that alternative world-modelling frameworks, like the recently proposed Synthetic Cognition~\cite{irga24} and JEPA~\cite{bpl23} algorithms, are better suited to avoid catastrophic forgetting and overfitting. The later uses neural networks to build modules, where each module has only one task, hence not suffering from catastrophic forgetting, while the former one is a totally new approach based on building abstract representations. These approaches are examples of proposals that have understood the limitations posed by the optimisation setup, and aim to switch to a world-modelling setup where optimisation is not the driving force of learning. Hence, they are promising avenues of research in the path to develop an artificial cognition.

Finally, this paper briefly discuss about other problems that the optimisation setup poses to solve the Artificial General Intelligence problem. Specifically, it discuss about its validity as a proxy for developing intelligence, about its potential for performing other tasks, about the plausibility of overfitting, and about the need for methods that say ``I do not know''. These discussions aim to give some context and further motivate researchers to use the current machine learning and deep learning methods (developed under this optimisation setup) for what they are valid, and discourage researchers to use them for what they cannot accomplish. They also aim to encourage researchers to look for alternatives, opening new avenues of research that could be more in line with what can be expected from an intelligence.

The rest of the paper is organised as follows: It presents a related work about the caveat in Section~\ref{sec:relwork}. Then, it defines optimisation methods in Section~\ref{sec:opt}. It explores the catastrophic forgetting caveat in Section~\ref{sec:forg}, and the overfitting problem in Section~\ref{sec:gen}. It briefly evaluates a recently proposed world-modelling framework in Section~\ref{sec:alt}. It discuss the results in Section~\ref{sec:disc}. Finally, it closes with the conclusions and future work in Section~\ref{sec:conc}.

\section{Related Work}\label{sec:relwork}
The Artificial Intelligence field started in $1956$ in a workshop~\cite{kp22}, and quickly focused into the development of so called ``expert systems''. However, these systems were produced rather manually, and required a lot of knowledge from its creators. Thus, a novel way of developing such systems was necessary. Here is where machine learning was proposed~\cite{kp22}. The idea was simple: let the method learn from the data. This led to the development of methods that use data to learn. However, learning was no easy task, and in the supervised learning field a solution quickly become the preferred option: let us learn by minimising the error produced when processing the training set. And thus the loss function became a necessary staple of any machine learning algorithm. Based on this function, new algorithms were developed, this time focused on solving the optimisation problem pose by the loss function. This is what one could call \emph{optimisation methods}.

These optimisation methods range from simple Logistic Regressions~\cite{cox58} to more complex Support Vector Machines~\cite{cv95}, and their common feature is that they guide their training by a loss function. With the development of these methods, a new, more powerful one arose that started a new field: the Artificial Neural Network (ANN)~\cite{mp90}. It started the deep learning field due to its versatility and performance, overcoming any other method. And developing the ANN to deal with sequences the Transformer~\cite{vspujgkp17} came to be.

The Transformer is an evolution of the ANN that has managed to obtain astonishing results, even beating the Turing Test~\cite{biever23}. However, even although a lot of effort and resources have been poured to develop the next step towards an artificial cognition, the current approach (training bigger and bigger Transformers) have arose minimal results, disappointing those that expected the next jump in intelligence. The closer to such a jump we have got are some reasoning-like behaviours that have been recently achieved using LLMs and textual chain of thoughts~\cite{ssfw+23,avllzy24,ygkgr24,rphb24}, but no reasoning mechanisms have arise~\cite{ckp23,hcmzysz24,smahbf25}. This paper aims to explain why this is the case, and why we need a different approach to develop the next step towards Artificial General Intelligence.

To address this problem, the paper focuses on a common issue of the aforementioned methods called \emph{catastrophic forgetting}. This issue arises when a method abruptly and drastically forgets previously learned information upon learning new information~\cite{mc89}. This is a well documented problem~\cite{pgak18,gpfb19,lspt21,mdt21,cceb22,zjwml22,sawant23,accf24,kalb24,ksrp25} with huge implications, because it incapacitates these methods to be used in continual learning or lifelong learning scenarios, impeding any kind of sequential leaning similar to what we humans do during all of our lives. As such, the scientific community has long try to solve this problem, with varying degrees of success. There are three main approaches to address this problem: ad-hoc methods, alternative learning procedures, and architectural approaches.

The ad-hoc methods usually try to ameliorate, mitigate, or control the impact of catastrophic forgetting using minor control measures. This includes methods like Elastic Weight Consolidation~\cite{vb19}, Synaptic Intelligence~\cite{zpg17}, Domain Shift~\cite{cz25} and others~\cite{bhmr22,zjsyxn24,zsttg24,gl25}. Their main goal is to allow some parameters to change, while others stay unmodified, so the original task can still be performed while new tasks can be learned. However, they manage to mitigate catastrophic forgetting but they do not avoid it.

The alternative learning procedures usually address catastrophic forgetting trough \emph{rehearsal}~\cite{robins93,amsr21,hcwylsys24,yyykr24}. Rehearsal is a technique that consist in storing some of the training samples of the initial training, and insert them between samples during new learning phases. This way, in the new learning phase, the model is not only learning the new task, but also remembering the old ones. Additional approaches use generative replay or similar rehearsal methods~\cite{pjpro25}. Finally, more recent approaches focus on modifying continuous learning and life-long learning in a rehearsal-like way~\cite{ka23,clhcqz24,hma25,lsldqc25,rs25,sgp25}. However, similar to the previous case, these methods only manage to mitigate the effects of catastrophic forgetting.

Finally, the last approach to addressing catastrophic forgetting consists on modifying the underlying architecture of an ANN. The most famous example of these approaches are Kolmogorov-Arnold Neural Networks (KAN)~\cite{lwvrhsht24}, where the neuron is transformed from a set of weights in the edges and an activation function in the node to a set of activation functions in the edges and an aggregation function in the node. However, although initially they seem to have solved the catastrophic forgetting problem present in traditional ANNs, recent research has discover that they still suffer catastrophic forgetting, but in a different way~\cite{lgzk25}. Another examples of architectural approaches include dynamically expanding networks~\cite{rrdskkph16,wly24}, selective architectures~\cite{fbbzhrpw17}, and attention-based mechanisms~\cite{ssmk18}, among others~\cite{brs24}. Their end goal is to isolate task-specific knowledge and reduce interference between tasks, but, similar to previous cases, they are unable to avoid catastrophic forgetting.

In the end, is this apparent stickiness of the catastrophic forgetting problem what motivates this paper: if all these different methods suffer from catastrophic forgetting, there should be a reason based on their common features. However, the bulk of the research in this field has tried to overcome catastrophic forgetting, without first trying to explain and understand its causes. Thus, the work presented in this paper tries to fill that proverbial hole.

\section{Optimisation Methods}\label{sec:opt}
To explore any caveat formally, first we need to define what are the objects we are dealing with. To that end, we need to define what is a solution to a problem. If we consider the process by which the inputs are transformed into outputs as a black box, we can argue that the solution to any problem is a mapping.

\bdfn
A mapping $M: I \to O$ is an association that associates each possible input $i \in I$ to a possible output $o \in O$.
That is, $\forall i\in I, \exists o\in O \| M(i) = o$,
where $I$ is the set of possible inputs, and $O$ is the set of possible outputs.
\edfn

Given the general definition of mapping, we can give a few more details about how the input is mapped into the output by knowing which weights were used when transforming the input into the output.

\bdfn\label{def:weimap}
A mapping $M_w: I \to_w O$ is a weighted mapping such that $\forall i\in I, \exists o\in O, w \in W \| M_w(i) = i\circ w = o$, where $\circ$ is any set of operations that use both $i$ and $w$ to produce $o$, and $W$ is the set of all possible weights.
\edfn

This more nuanced mapping will be useful when defining catastrophic forgetting. Now, let us define what is an optimisation problem. In mathematical terms, an optimisation problem is usually defined as follows.

\bdfn
An optimisation problem $Q$ is defined as:
$$minimise_x\ f(x) \| \left \{
\begin{array}{rl}
    g_i(x) \leq 0, & i = 1,\dots,m\\
    h_j(x) = 0, & j=1,\dots,p
\end{array}
\right .$$

Where $f(x):\mathbb{R}^n \to \mathbb{R}$ is the objective function to be minimised, $g_i(x)$ are the inequality constraints, $h_j(x)$ are the equality constraints, $m \geq 0$, and $p \geq 0$.
\edfn

Now, from a machine learning perspective, $f(x)$ is the loss function for a set of predicted outputs $y_{pred} \in O$ compared with a set of expected outputs $y_{true} \in O$. In fact, we do not have inequality constraints, neither equality ones, that is, $m = 0$ and $p = 0$. Thus, we can simplify the machine learning optimisation problem to be as follows.

\bdfn
A machine learning optimisation problem $Q$ is defined as:
$$minimise_x\ f_{y_{true}}(x)$$

Where $f_{y_{true}}(x):\mathbb{R}^n \to \mathbb{R}$ is the objective function to be minimised, that depends on the expected outputs $y_{true} \in O$.
\edfn

The rest of the paper will use this definition of optimisation problem. Here, it is critical to remark that $f(x)$ is the loss function, and as such the set of expected values $y_{true}$ is fixed, and the set of predicted values $y_{pred}$ is its input. That is, $x = y_{pred}$. The underlying relationship between $y_{true}$ and $y_{pred}$ is that they are respectively the true and predicted values corresponding to an specific set of inputs. Thus, the goal of most machine learning algorithms is to build $y_{pred}$ as similar as possible to $y_{true}$ based on the data of the corresponding inputs.

Given this, a solution to any optimisation problem has the shape of a mapping $M: I \to y_{true}$ from inputs to their corresponding values of $y_{true}$. However, as the goal is to predict $y_{true}$ from the inputs without knowing the actual $y_{true}$ values, then the proposed solutions have the shape of a mapping $M: I \to O$, with $M(I) = y_{pred}$.

Now, if the solution provided by an optimisation method is a weighted mapping $M_w: I \to_w O$ with $M_w(I) = y_{pred}$, we can explore what is the relationship between such mapping and its optimisation problem. Let us start by stating that two mappings for two different optimisation problems would produce different results.

\bthm\label{thm:diffQ}
Given an input set $I$ and an output set $O$, a training set $L \in I$ that produces an optimisation problem $Q = minimise_x\ f_{y_{true}}(x)$ with solution $M_w: I \to_w O$, and another training set $L' \in I$, such that $L \cap L' = \emptyset$, that produces a different optimisation problem $Q' = minimise_x\ f_{y'_{true}}(x) \neq Q$ with solution $M_{w'}: I \to_w' O$, then $M_w(L) = M_{w'}(L)$ and $M_w(L') = M_{w'}(L')$ if and only if $w = w'$.
\ethm
\begin{proof}
    Let us assume that $M_w(L) = M_{w'}(L)$ and $M_w(L') = M_{w'}(L')$ but $w \neq w'$. Then, by Definition~\ref{def:weimap}, $M_w \neq M_{w'}$. Now, as $M_w$ solves $Q$, and $M_{w'}$ solves $Q'$, with $Q \neq Q'$, then $i\circ w \neq i\circ w'\ \forall i \in I$, where $\circ$ is the operations performed by the mapping $M$. This means that, $M_w(L) \neq M_{w'}(L)$ and $M_w(L') \neq M_{w'}(L')$, what is a contradiction.

    Now, let us assume that $w = w'$ but $M_w(L) \neq M_{w'}(L)$ and $M_w(L') \neq M_{w'}(L')$. Then, by Definition~\ref{def:weimap}, $i\circ w \neq i\circ w'\ \forall i \in L \cup L'$, and thus $M_w \neq M_{w'}$. Then, $w \neq w'$, what is a contradiction.
\end{proof}

This is a very intuitive result, as two problems have two different solutions. However, that does not makes it less important, as this will be crucial later on. Now, let us explore what happens in the case that the two mappings solve the same optimisation problem.

\bthm\label{thm:sameQ}
Given an input set $I$ and an output set $O$, a training set $L \in I$ that produces an optimisation problem $Q = minimise_x\ f_{y_{true}}(x)$ with solution $M_w: I \to_w O$, and another training set $L' \in I$, such that $L \cap L' = \emptyset$, that produces a different optimisation problem $Q' = minimise_x\ f_{y'_{true}}(x) \equiv Q$ with solution $M_{w'}: I \to_{w'} O$, then $M_w(L) = M_{w'}(L)$ and $M_w(L') = M_{w'}(L')$.
\ethm
\begin{proof}
    Let us start by the trivial case: if $w = w'$ then $M_w = M_{w'}$, and thus $M_w(L) = M_{w'}(L)$ and $M_w(L') = M_{w'}(L')$.

    Now, let us assume that $w \neq w'$. Then, by Definition~\ref{def:weimap}, $M_w \neq M_{w'}$. Now, as $M_w$ solves $Q$, and $M_{w'}$ solves $Q'$, with $Q \equiv Q'$, then $i\circ w = i\circ w'\ \forall i \in I$, where $\circ$ is the operations performed by the mapping $M$. This means that, $M_w(L) = M_{w'}(L)$ and $M_w(L') = M_{w'}(L')$.
\end{proof}

As a conclusion of this section, we can define an optimisation method as a process that aims to build a mapping $M: I \to O$ that minimises the loss function $f(x)$. The rest of the paper will use this definition when talking about optimisation methods.

\bdfn
Given an input set $I$, an output set $O$, and a training set $L \in I$ that produces an optimisation problem $Q = minimise_x\ f_{y_{true}}(x)$, we define an \emph{optimisation method} as a process $P$ that produces a mapping $M: I \to O$ with $M(I) = y_{pred}$, such that $y_{pred}$ minimises the loss function $f$. That is, $M$ solves $Q$.

The mapping $M$ is produced by $P$ after learning $L$, with $M(L) = y_{L}$.
\edfn

\section{Catastrophic Forgetting}\label{sec:forg}
Catastrophic forgetting is a problem that is widely associated to Artificial Neural Networks, but that is not exclusive of them. It was first detected by McCloskey and Cohen in 1989~\cite{mc89}, and has been a huge problem for machine learning. The original definition poses the problem as the fact that the method abruptly and drastically forgets previously learned information upon learning new information~\cite{mc89}. In mathematical terms, we can redefine it for our mapping $M: I \to O$ as a change in previously learned assignments through new updates.

\bdfn
Given a process $P$ that produces a mapping $M: I \to O$ from an input set $I$ to an output set $O$, and given the set of learned inputs $L \in I$ such that $M(L) = y_{L}$, we define catastrophic forgetting as the fact that $P$, after trained with a new set of training inputs $L' \in I$ such that $L \cap L' = \emptyset$, produced a new mapping $M': I \to O$ such that $M'(L) = y'_{L}$ and $y_{L} \neq y'_{L}$.
\edfn

As can be deduced from this definition, not any method that produces a mapping would suffer from catastrophic forgetting. For example, a method that produces a table (like the initial versions of reinforcement learning) would not suffer catastrophic forgetting, because when learning from new inputs (i.e. new states), those new inputs would create new table entries and not modify the previously existing ones. However, a key difference here is that using a table as knowledge representation does not really produce a mapping $M: I \to O$ but instead a mapping $M: L \to O$ where $L \in I$ is the subset of learned inputs. This is a very literal approach that has little usability, and most methods try to generalise in some way with the goal of producing a mapping $M: I \to O$. And it is this generalisation the one that motivated methods that suffer from catastrophic forgetting, because the main method to obtain such generalisation is defining $M$ as a weighted mapping $M_w: I \to_w O$.

\bthm
Given an input set $I$, an output set $O$, a training set $L \in I$ that produces an optimisation problem $Q$, and an optimisation method $P$ trained with $L$ that produces a weighted mapping $M_w: I \to_w O$. If we have a new training set $L' \in I$ that produces a new optimisation problem $Q'\neq Q$ such that $L \cap L' = \emptyset$, and we train $P$ with $L'$, then $P$ will suffer from catastrophic forgetting.
\ethm
\begin{proof}
    Let us assume that $M_w(L) = y_L$, $M_w(L') = y_{L'}$, and that $P$ produced a new weighted mapping $M_{w'}: I \to_{w'} O$ after learning $L'$ such that $M_{w'}(L') = y'_{L'}$ and $M_{w'}(L) = y'_L$. Now, we can assume, without loss of generality, that $y_{L'} \neq y'_{L'}$ (if $y_{L'} = y'_{L'}$ we can assume no learning has taken place). That implies that $w \neq w'$ because $i \circ w \neq i \circ w'\ \forall i \in L'$, and thus $M_w \neq M_{w'}$, with $M_w$ solving $Q$ and $M_{w'}$ solving $Q' \neq Q$.
    
    Now, let us assume that $y_L = y'_L$, then that implies that $M_w(L) = M_{w'}(L)$, what can be possible if and only if $w = w'$ by Theorem~\ref{thm:diffQ}. However, this is a contradiction. Thus, $y_L \neq y'_L$, what means that $P$ suffered catastrophic forgetting.
\end{proof}

This is a fundamental result, because it proves that any optimisation method based on weights will suffer catastrophic forgetting when trained for new tasks. Now, let us explore some cases where the update of the weights is not a problem, for example, when the training inputs correspond to the same problem.

\bthm
Given an input set $I$, an output set $O$, a training set $L \in I$ that produces an optimisation problem $Q$, and an optimisation method $P$ trained with $L$ that produces a weighted mapping $M_w: I \to_w O$. If we have a new training set $L' \in I$ that produces the same optimisation problem $Q' \equiv Q$, and we train $P$ with $L'$, then $P$ will not suffer from catastrophic forgetting.
\ethm
\begin{proof}
    Let us assume that $M_w(L) = y_L$, $M_w(L') = y_{L'}$, and that $P$ produced a new weighted mapping $M_{w'}: I \to_{w'} O$ after learning $L'$ such that $M_{w'}(L') = y'_{L'}$ and $M_{w'}(L) = y'_L$. Now, we can assume, without loss of generality, that $y_{L'} \neq y'_{L'}$ (if $y_{L'} = y'_{L'}$ we can assume no learning has taken place). That implies that $w \neq w'$ because $i \circ w \neq i \circ w'\ \forall i \in L'$, and thus $M_w \neq M_{w'}$, with $M_w$ solving $Q$ and $M_{w'}$ solving $Q' \equiv Q$.
    
    Now, let us assume that $y_L \neq y'_L$, then that implies that $M_w(L) \neq M_{w'}(L)$ and $w \neq w'$. However, as $Q \equiv Q'$, then, by Theorem~\ref{thm:sameQ}, $w'$ are weights such that $i \circ w' = i \circ w\ \forall i \in L\cup L'$. Thus, $y_L = y'_L$, what means that $P$ did not suffered catastrophic forgetting.
\end{proof}

This result shows that, when the underlying optimisation problem is the same, training with new inputs is not a problem. Actually, this explains why we do not suffer catastrophic forgetting when training with new inputs from the same problem. Moreover, this case is fundamental because without it, there would be no learning.

To end this section, it is important to remark that this problem affects to any method whose learning is guided by a loss function (either explicit or implicit), that is, to any optimisation method. In particular, Artificial Neural Networks and their derivatives (like Transformers) are trained in this way, and thus are limited by this problem. Thus, a single, huge neural network will be unable to solve multiple, different tasks. However, that does not impede a set of interconnected optimisation methods, each one focused on a single task, to solve multiple different tasks, as long as each method is solving only one, specific task. That is the aim of proposals like Meta's JEPA~\cite{bpl23}. Exploring this scenario scapes the scope of this paper, but similar reasoning can be applied to any other method guided by a loss function to minimise.

\section{Overfitting}\label{sec:gen}
Generalisation is the holy grail of the Artificial Intelligence field. It allows any method to give an answer for inputs not previously seen, and thus to solve problems for more inputs than the already learned ones. Therefore, generalisation is the capability to provide a correct answer for inputs outside of the training input set. However, optimisation methods tend to generalise poorly when they have focused too much on the training set. This effect is what is called \emph{overfitting}, and greatly hampers the generalisation capabilities of optimisation methods.

The main problem with overfitting is that it makes solving the optimisation problem useless. Let me put it this way: we have methods specifically designed to solve an optimisation problem, but the better they solve the problem, the less useful its result is for us because it generalises poorly. Therefore, we need to approximate the problem but not solve it. Thus, we need to hamper our methods so they are not so good solving the optimisation problem. And then we have to find the ``right balance'' between solving the optimisation problem and not solving it enough.

Clearly, this is not an ideal scenario. However, this is fundamentally inherent to any optimisation method. To prove it, let us start by defining what is overfitting.
\bdfn
Given a process $P$ that produces a mapping $M: I \to O$ from an input set $I$ to an output set $O$, and given the set of learned inputs $L \in I$ that produces an optimisation problem $Q = minimise_x\ f_{y_{true}}(x)$ such that $M(L) = y_{L}$, we define overfitting as the fact that, for any new set of inputs $L' \in I$ such that $L \cap L' = \emptyset$ that produces an optimisation problem $Q' = minimise_x\ f_{y'_{true}}(x) \approx Q$, $P$ produces a sub-mapping $M(L') = y'_{L'}$ such that $y'_{L'}$ does not minimise $f_{y'_{true}}(L')$, that is, it does not solve $Q'$.
\edfn

With this definition, we can explore what happens when we have an optimisation method.

\bthm
Given an input set $I$, an output set $O$, a training set $L \in I$ that produces an optimisation problem $Q = minimise_x\ f_{y_{true}}(x)$, and an optimisation method $P$ trained with $L$ that produces a weighted mapping $M_w: I \to_w O$. If we have a new set of inputs $L' \in I$ that produces a new optimisation problem $Q' = minimise_x\ f_{y'_{true}}(x) \approx Q$ such that $L \cap L' = \emptyset$, then $P$ will suffer from overfitting.
\ethm
\begin{proof}
    Let us assume that $P$ solves $Q$ and $P$ does not suffer from overfitting. Then, $P$ solves $Q'$, that is, $M(L') = y'_{L'}$ minimises $f_{y'_{true}}(L')$. However, $Q'\approx Q$ but $Q'\neq Q$, and thus, by Theorem~\ref{thm:diffQ}, $P$ cannot solve $Q$ too, what is a contradiction.
\end{proof}

This result is crucial to explain why we always obtain better train accuracies than test accuracies: because the optimisation function for an input set, although it is very similar to the one for the training set, it is still slightly different. Thus, the closer we are to solve the training minimisation problem, the further away we are to solve the test minimisation problem, and hence the balance we need to achieve.
This result also explains why machine learning (and specially deep learning) methods suffer so much from overfitting: because they are weight-based algorithms that produce a mapping based on a set of weights.

Here, it is important to remark that there is plenty research into mitigation methods for overfitting, like regularisation~\cite{bengio12}, dropout~\cite{shkss14}, and data augmentation~\cite{ksh12,sk19}. However, these methods only mitigate the effects of overfitting and delay their appearance, but they have been unable to avoid it. The proof shown in this section shows why they have been unable to solve this problem: due to the optimisation setup.

\section{Exploring World-Modelling Alternatives}\label{sec:alt}
In order to make a positive contribution, this paper also explores how world-modelling alternatives do not suffer from either catastrophic forgetting nor overfitting. Hence, the goal is to show that promising avenues are open for any researcher aiming to contribute to the development of an Artificial General Intelligence algorithm. The key of these world-modelling methods is that they do not approach the intelligence problem as an optimisation one, but instead they approach it as a modelling problem: the end goal is to model the world from the received perceptions, and thus give order and ``meaning'' to it. This approach allows these methods to build non-weighted mappings as solutions, and hence they can avoid the problems discussed in previous sections.

Here, it is important to remark that, as detailed in Section~\ref{sec:relwork}, there is plenty of empirical evidence of optimisation methods suffering from catastrophic forgetting and overfitting. However, there is little evidence of other kind of methods not suffering these problems. Hence, an empirical research about the capability of alternative methods to avoid catastrophic forgetting and overfitting has its own value. The only limitation impose to those alternative methods would be that they should be able to obtain similar results to those obtained by the optimisation methods. Given the nature of optimisation methods, it would be difficult to obtain better results, but that cannot justify obtaining bad results, because then those alternative methods are not a real alternative to the current state-of-the-art.

Recently, a novel framework to develop Artificial Intelligence algorithms have been proposed, called Synthetic Cognition~\cite{irga24}. This framework develops a representation-based model whose focus is to model the input domain, and then, thanks to its primitive-based mechanism, it is able to use such model to produce answers for each input. Thus, this framework does not suffer from the optimisation problem. This framework has been implemented into an algorithm called Unsupervised Cognition~\cite{iarga24}, that has obtained astounding results, not only getting similar results to those obtained by optimisation methods, but actually becoming the state-of-the-art in some cases~\cite{iarga24,irarg25}. This algorithm is a world-modelling algorithm that does not suffer from the optimisation problem discussed in this paper, thus it is a promising avenue for developing alternative Artificial Intelligence algorithms. To explore how this framework does not suffers from catastrophic forgetting nor overfitting, two experiments were performed: a catastrophic forgetting one and an overfitting one.

\subsection{Catastrophic Forgetting Experiment}
The catastrophic forgetting experiment consisted on instantiating an Unsupervised Cognition, and train it for two different tasks. In this case the tasks were to solve the Wisconsin Breast Cancer~\cite{wsm95, dg17} and Pima Indians Diabetes~\cite{sedkj88} datasets. These are two purely numerical datasets, and thus they have more probability of producing overlapping representations than, let us say, MNIST and Wisconsin Breast Cancer. The instance of Unsupervised Cognition was trained first in the Wisconsin Breast Cancer dataset, evaluated over a test set (obtaining a $95.32\%$ of accuracy), then trained over the Pima Indians Diabetes dataset (obtaining a $69.70\%$ accuracy), and then evaluated again over the same test set for the Wisconsin Breast Cancer dataset, obtaining the same accuracy of $95.32\%$.

To reinforce the previous results, the experiment was done also in the other direction: a new instance of Unsupervised Cognition was trained first with the Pima Indians Diabetes dataset, evaluated over a test set (obtaining a $69.70\%$ accuracy), then trained over the  Wisconsin Breast Cancer dataset (obtaining a $95.32\%$ of accuracy), and then evaluated again over the same test set for the Pima Indians Diabetes dataset, obtaining the same accuracy of $69.70\%$. These experiments empirically prove that this approach does not suffer of catastrophic forgetting, independently of the training order.

These results should be no surprise to anyone familiarised with the aforementioned framework. Synthetic Cognition is a framework based on building representations of the inputs. Thus, it has a representation for each input, as well as aggregative abstractions that help the framework to generalise. These representations are the key to its protection from catastrophic forgetting: as each problem is different, it builds different representations that do not overlap with each other. Thus, in its internal structure, there are in fact different representation hierarchies for each problem, thus impeding the modification of previous representations during the second learning phase. Hence. the learning performed to solve the first problem is intact and can still be used to solve that problem.

This experiment is quite limited in scope, and thus does not account for all the possible variability and overlap that two tasks can have when building representations. However, it is a first proof that these kind of methods start from a better position to avoid catastrophic forgetting. Further research is necessary to explore if there are corner cases where these methods can suffer catastrophic forgetting, or if they are totally immune. However, those experiments fall out of the scope of this paper.

\subsection{Overfitting Experiment}
The overfitting experiment consisted on instantiating an Unsupervised Cognition, and training it to solve a given task multiple times, trying to force the appearance of overfitting. In this case, the task to solve was the Wisconsin Breast Cancer dataset previously used. To evaluate the appearance of overfitting, both the accuracy and the internal representations where monitored. The goal was to detect not only when the overfitting was reducing the accuracy, but also when the learning was not updating the internal representations, and thus when overfitting can not happen due to changes not being produced internally.

In this case, the learning stopped producing changes into the internal representations of the Unsupervised Cognition instance at the $5$th epoch, as this is a quite simple dataset. However, just to be sure, the algorithm was trained $10$ times with the same training dataset, and evaluated with a separate test dataset. The evaluation arose consistently a $95.32\%$ accuracy for all epochs, but internal changes where produced during the $5$ first epochs. Thus, although minimal changes where produced internally, it is clear that more epochs did not lead to a loss in test accuracy due to overfitting.

This result should be no surprise to anyone familiarised with the aforementioned framework. Synthetic Cognition is a framework based on building representations of the inputs, and thus it has a literal representation of each input, as well as additional representations (built by an aggregative abstraction mechanism) that allow the algorithm to generalise. Thus, subsequent epochs after the first one only slightly modify these additional representations, until an stable representation is found. Once this stable representation is found, there is no further modification to these additional representations, and thus subsequent epochs have no effect at all. Hence, it is very difficult for phenomena like overfitting to appear in this kind of solutions.

This experiment is quite limited in scope, and thus does not account for all the possible ways such small modifications to the additional representations can produce overfitting-like effects. However, it is a first proof that these kind of methods start from a better position to avoid overfitting. Further research is necessary to explore if there are corner cases where these methods can suffer overfitting, or if they are totally immune. However, those experiments fall out of the scope of this paper.

\section{Discussion}\label{sec:disc}
This section discusses several facts about the optimisation problem setup analysed in this paper. Specifically, about its validity as a proxy for developing intelligence, about its potential for performing other tasks, about the plausibility of overfitting, and about the need for methods that say ``I do not know''.

The first thing to discuss is how ill posed is the optimisation problem as a proxy for developing intelligence. We observe that, as defined before, an optimisation problem aims to build a mapping $M: I \to O$ with the optimal solution. However, a cognitive agent, one able to reason, does not always give the optimal solution. A key factor for that is that a cognitive agent needs to be able to ponder all the possible solutions and then give the one it \emph{thinks} correct. To that end, such agent should be able to have a representation of all possible solutions to be able to reason about them. And the optimisation problem approach hides all of this and only asks for the optimal solution, irrespective of how such solution has been obtained. This opens the door for a lot of approaches that give the optimal solution, but that do not build the tools to reason, what one could argue has happened in the machine learning field, specially with the deep learning field.

Secondly, it is important to remark that the setup discussed in this paper is not inherently bad. For example, generative models like diffusion models take advantage of this setup in order to generate novel outcomes, and there is a huge literature exploiting the properties of the setup. Also, for solving a specific task, the setup is still useful, as any method trained for a specific task will not suffer from catastrophic forgetting. This paper just signals how this setup has some limitations to develop Artificial General Intelligence in order to right the course of research in that area, but it is nonetheless important to promote the research into models that use the aforementioned setup for the tasks they are suited for.

Additionally, it is important to discuss how overfitting is cognitively implausible. First, it is important to remark that it is derived from the optimisation setup, where the problem surges from learning something ``too much''. However, for cognitive agents, learning something ``too much'' as to forget how to properly recognise objects is not a thing. The cognitive equivalent of overfitting would be a human that is unable to recognise a table in a store because it has seen too many times its home and work tables. That is something unheard of, and thus it is clearly not a property of cognitive agents.

Finally, the last topic to discuss is why the capability of saying ``I do not know'' is fundamental for any method aiming to develop Artificial General Intelligence. A problem of the optimisation problem setup is that it requires the produced mapping $M: I \to O$ to provide an answer for any possible input, what produces undesirable side effects. Due to the finite amount of data, it is impossible to define a mapping for every possible input with high reliability, and thus there will be areas for which the answer will be provided with low reliability. This in fact produces weird situations, like a method that learned to distinguish between cats and dogs saying that a photo of a car is a cat; but it also produces dangerous flaws, like the weakness against adversarial attacks. Using a different approach that allows the answer to be ``I do not know'' would be key to avoid this kind of undesirable side effects.

\section{Conclusions}\label{sec:conc}
Artificial Intelligence has always focused on building intelligence by building methods that produce similar effects to those produced by a natural intelligence (i.e. an human). However, this approach has lead to the development of optimisation methods that do not ensure that such ``intelligence effects'' are produced in an intelligent way. In fact, such methods typically suffer from catastrophic forgetting, a huge deterrent to build truly cognitive agents.

This paper has proved that optimisation methods cannot avoid catastrophic forgetting, and thus they are not the way towards solving the Artificial General Intelligence problem. Moreover, it has discussed about how overfitting is also inherent to these methods and how it also impedes the development of Artificial General Intelligence. Additionally, it has shown how alternative methods like world-modelling ones can overcome these problems. Finally, it has discussed other, smaller problems posed by optimisation methods, and disclaimed that this does not mean these methods cannot be used for other purposes, like generative methods.

As a general conclusion, it is important to remark that, in the path towards Artificial General Intelligence, we have to divert from the weight-based optimisation methods, and look elsewhere. A promising avenue of research has been open by modelling methods~\cite{irga24} and modular architectures~\cite{bpl23}, but exploring them is matter for future work. Here, the paper just reflects on the fact that, even although weight-based optimisation methods have become very popular due to their astonishing results, we need something else to develop Artificial General Intelligence. To develop these alternatives we need not only framing the problem as a non-optimisation one, but also defining new mental frameworks about how to address the Artificial General Intelligence problem, like the recently published Synthetic Cognition one~\cite{irga24}. Finally, it is critical to explicit that optimisation tasks are still valid evaluation of cognitive performance. Here it has been proved only that such framework is not enough to develop good solutions. However, any other framework should produce solutions that solve these optimisation tasks anyway.

As future work, it would be relevant to perform a deeper analysis of the capabilities of world-modelling methods. It would be also interesting to explore if these problems extrapolate to sets of optimisation methods. Finally, it would be critical to propose alternative methods to the already established ones, to start developing non-optimisation methods.

\section*{Acknowledgments}
I want to thank my team at Avatar Cognition for our insightful discussions about the topic and the support to write this paper.
This work has been supported by the Torres-Quevedo grant PTQ2023-012986 funded by the MCIU/AEI /10.13039/501100011033.

\section*{Declarations of Conflict of Interest}
The authors declared that they have no conflicts of interest to this work.

\bibliographystyle{plain}
\bibliography{biblio}

\vfill


\begin{IEEEbiographynophoto}{Alfredo Ibias}
is the Lead Researcher at Avatar Cognition at Barcelona, Spain. His research interests include Artificial General Intelligence, Robotics, and Software Testing. Ibias received his Ph.D. in Computer Science from Universidad Complutense de Madrid. He is a member at IEEE and ACM. Contact him at alfredoibias.com.
\end{IEEEbiographynophoto}

\vfill

\end{document}